\documentclass{article}
\usepackage{spconf,amsmath,graphicx}
\usepackage{color}
\usepackage{soul}
\usepackage{booktabs}
\usepackage{multirow}
\usepackage{subfigure}
\usepackage{xspace}

\newcommand{\sysname}{\texttt{CAT}\xspace}
\usepackage{algpseudocode}
\usepackage{algorithm}
\usepackage{amssymb}
\usepackage{algpseudocode}
\floatname{algorithm}{Protocol}

\title{Case-Aware Adversarial Training}
\name{Mingyuan~Fan$^1$,~Yang~Liu$^2$,~Cen~Chen$^1$}
\address{$^1$School of Data Science and Engineering, East China Normal University, China\\
$^2$School of Cyber Engineering, Xidian University, China
}
\begin{document}
%
\maketitle
\begin{abstract}
The neural network (NN) becomes one of the most heated type of models in various signal processing applications. However, NNs are extremely vulnerable to adversarial examples (AEs). To defend AEs, adversarial training (AT) is believed to be the most effective method while due to the intensive computation, AT is limited to be applied in resource-limited applications. In this paper, to resolve the problem, we design a generic and efficient AT improvement scheme, namely case-aware adversarial training (CAT). Specifically, the intuition stems from the fact that a very limited part of informative samples can contribute to most of model performance. Alternatively, if only the most informative AEs are used in AT, we can lower the computation complexity of AT significantly as maintaining the defense effect. To achieve this, CAT  achieves two breakthroughs. First, a method to estimate the information degree of adversarial examples is proposed for AE filtering. Second, to further enrich the information that the NN can obtain from AEs, CAT involves a weight estimation and class-level balancing based sampling strategy to increase the diversity of AT at each iteration. Extensive experiments show that CAT is faster than vanilla AT by up to 3x while achieving competitive defense effect.
\end{abstract}

\section{Introduction}
Over the past decade, the neural networks (NNs) has become the most outstanding approach in a wide range of signal processing tasks due to its remarkable data processing ability.
However, despite the promising performance, NN is extremely vulnerable to the attack of adversarial examples~\cite{PID,bayes_attack,surfree,non_gradient_attack,auto_pgd}.
By adding human-imperceptible and carefully crafted tiny noises, adversarial examples can trick the victim model to behave as desired by the attacker~\cite{adv_to_med,adv_healthcare,adv_for_obj,adv_for_selfdriving2}.
Due to the great security threat caused by adversarial examples, an effective adversarial defense method becomes growingly much-needed in applications.

Currently, three methods are mainly explored to achieve adversarial defense, including detection~\cite{detect_autoencoder}, data preprocessing~\cite{defense_gan} and adversarial training (AT)~\cite{pgd_training}.
Among them, detection aims to reveal the malicious inputs mixed in the normal ones and avert the attack before it happens.
Data preprocessing achieves defense by transforming malicious inputs into normal inputs.
However, both of these two kinds of defense are proved to be breakable in~\cite{obfuscated_grad}.
At last, AT becomes the only way believed to be able to raise the robustness of NNs against adversarial examples intrinsically~\cite{bag_of_tricks}.


AT achieves defense in a straightforward manner, i.e., making the model trainer to actively add adversarial examples into the training set to induce the model to learn how to correctly classify them.
However, due to the introduction of the generation of adversarial examples, the cost of the AT can be dozens of times more compared with normal training.
Considering the long time used for training a NN, such as training a VGG-16 network with ImageNet, the applicable range of the naive AT method~\cite{pgd_training} is very limited.

In this paper, we propose a novel method to accelerate the AT process, called case-aware adversarial training (CAT).
The design of \sysname is first motivated by the fact revealed in the active learning technique~\cite{active_learning,deep_active_learning} that the decision boundaries of a NN model can be strongly affected by a rare number of training examples with rich information.
Thus, instead of using adversarial examples transformed from all training samples, a more efficient way to achieve AT is to actively select some informative adversarial examples and only use them in the AT process.
However, straightforward as the idea is, it cannot be effortlessly applied in practice because the information gain of adversarial examples cannot be derived from the original examples directly.
Alternatively, to select informative adversarial examples, we have to craft every sample into adversarial examples, which is still computation-intensive.

To get over the drawback, \sysname mainly achieve two breakthroughs.
First, we observe that adversarial examples of a certain example over successive iterations exhibit good similarity.
It suggests that adversarial examples crafted before can be employed to approximately estimate the importance of the example, without crafting its adversarial examples from scratch so that the cost of AT can be greatly reduced.
Second, even if each sampled example contains rich information, these examples may be quite similar, decreasing the overall effective information contained in the mini-batch.
To alleviate the issue, \sysname leverages two measures for increasing the diversity of the examples in the mini-batch, i.e., weighted sampling without replacement and class-level balancing.
Moreover, we highlight that \sysname is indeed orthogonal with other AT techniques, and it can be easily integrated into other frameworks of AT.

\textbf{Our contributions} can be summaries as follows:
\begin{itemize}
    \item We propose a novel scheme to accelerate AT, which can be used against adversarial examples. To this end, we design a weighted sampling strategy that involves a information gain estimation method to evaluate the importance of an adversarial example to adversarial training.
    
    \item We discover that during AT, the adversarial examples perform similarly between iterations. Based on the discovery, we can improve the weighted sampling process by deriving the gain of each adversarial example from its previous counterparts.

    
    \item We conduct extensive experiments to examine the effectiveness of the proposed scheme, and the experiment results show that CAT is faster than the conventional adversarial training scheme by up to 3x with competitive performance.
\end{itemize}


\label{intro}

\section{Approach}
\subsection{Problem Formulation}
This paper is designed to leverage adversarial training to achieve defense.
Formally, given the training set $D=\{(x_1,y_1), \cdots, (x_n,y_n)\}$ and an NN $F_{\theta}$, the objective of traditional adversarial training can be formulated as follows.
\begin{equation}\label{eq_at}
\min_{\theta} \sum_{i=1}^{n} \max_{\delta_i} L(F_{\theta}(x_i +\delta_i),y_i), 
~s.t.,~||\delta_i||_{\infty} \leq \epsilon,
\end{equation}
where $||\cdot||_{\infty}$ denotes the $\infty$-norm of inputs and $L(\cdot,\cdot)$ indicates the loss function (e.g., cross-entropy loss).
Besides, the noises $\delta_i$ used to synthesize adversarial examples are generated by Eq.~\ref{eq_noise}.
\begin{equation}\label{eq_noise}
\begin{aligned}
    \delta_i^{j} &= Proj_{\epsilon}(\delta_i^{j-1} +  \nabla_{\delta_i^{j-1}}L(F_{\theta}(x_i+\delta_i^{j-1}),y_i)),\\
\delta_i^{0} &= 0~,\delta_i = \delta_i^{m},~j = 1,2,\cdots,m,
\end{aligned}
\end{equation}
where $m$ is the given total iteration number, $Proj_{\epsilon}(\cdot)$ can project the inputs into $\epsilon$-balls \cite{PGD}, $\delta_i^{j}$ denotes the crafted adversarial noise in the j-th iteration, and $L(\cdot,\cdot)$ indicates the loss function (e.g., cross-entropy loss).
From Eq.~\ref{eq_at}, it can be observed that adversarial training enhances the robustness of $F_{\theta}$ against adversarial examples by making it to remember how to classify these adversarial examples correctly in the training stage.
However, since the past adversarial training schemes~\cite{pgd_training,channel_training} require all training samples to generate adversarial examples, their costs can be tens of times more than the normal training, which is totally intolerable for applications.
To break the bottleneck, we propose the following \sysname scheme.



\subsection{Case-aware Adversarial Training}
Compared with traditional schemes~\cite{pgd_training}, the basic improvement of \sysname (Algorithm~\ref{algo_CAT}) is the involvement of a novel sampling strategy that can select informative adversarial examples to maximize the gains of the model at each iteration.
Specifically, the sampling strategy is mainly composed of two parts: 1) measuring samples with the highest information gains; 2) maximizing the diversity of sample at each each training itearation.


\begin{algorithm}[ht!]
  \caption{CAT}
  \label{algo_CAT}
  \begin{algorithmic}[1]
    \Require $F$: the neural network; $N$: the number of iterations; $D$: the training dataset.

    \Ensure $F_{\theta_{1}}$: adversarially trained neural network.
    
    \State Initialize the weight of each sample in $D$ to be $w^i_0 = 1,~i=1,2,\cdots,n$.

    \For{$t \gets 1$ to $N$}
      \State Sample a batch of data $\mathbb{X},\mathbb{Y}$ from $D$ based on the information gains computed by Eq.~\ref{weight1}.
      \State Craft adversarial examples $\mathbb{X}_{adv}$ for $\mathbb{X}$.
      \State Update the parameters $\theta_i$ of $F$ with $\mathbb{X}_{adv},\mathbb{Y}$.
      \State Update the weight of $\mathbb{X}$ with $\mathbb{X}_{adv}$ based on Eq.~\ref{weight2}.
    \EndFor
    \State \textbf{Return}~~$F_{\theta_{N}}$.
  \end{algorithmic}
\end{algorithm}

\textbf{Measuring the information gain.}
Instead of using the adversarial examples crafted from every $x_i\in D$ as defined in Eq.~\ref{eq_at}, \sysname only uses parts of adversarial examples with high information gains during the training process to save the computation cost. 
Here, the information gain $w^i$ for each adversarial example $x_i +\delta_i$ is measured according to the fact that the higher the uncertainty of a sample is for a model, the more the model can gain from the sample (see uncertainty sampling~\cite{sampling_strategy}.
Based on the idea, $w^i$ can be computed as follows.
\begin{equation}
\label{weight1}
w^i = \max_{k \neq y_i,k=1,\cdots,K} {log(F_{\theta}(x_i+\delta_i)[k]}) - log(F_{\theta}(x_i+\delta_i)[y_i]),
\end{equation}
where $K$ is the number of classes, $log(F_{\theta}(\cdot)[k])$ denotes the likelihood of the $k$-th class, and $y_i$ is the ground-truth label.
Intuitively, Eq.~\ref{weight1} describes the uncertainty of $x_i+\delta_i$ by computing its maximum likelihood distance.
The higher $w^i$ indicates the lower confidence for $F_\theta$ to correctly classify $x_i+\delta_i$ (higher probability to classify the adversarial example into another class).


\begin{figure}
    \centering
    \includegraphics[width=0.6\linewidth]{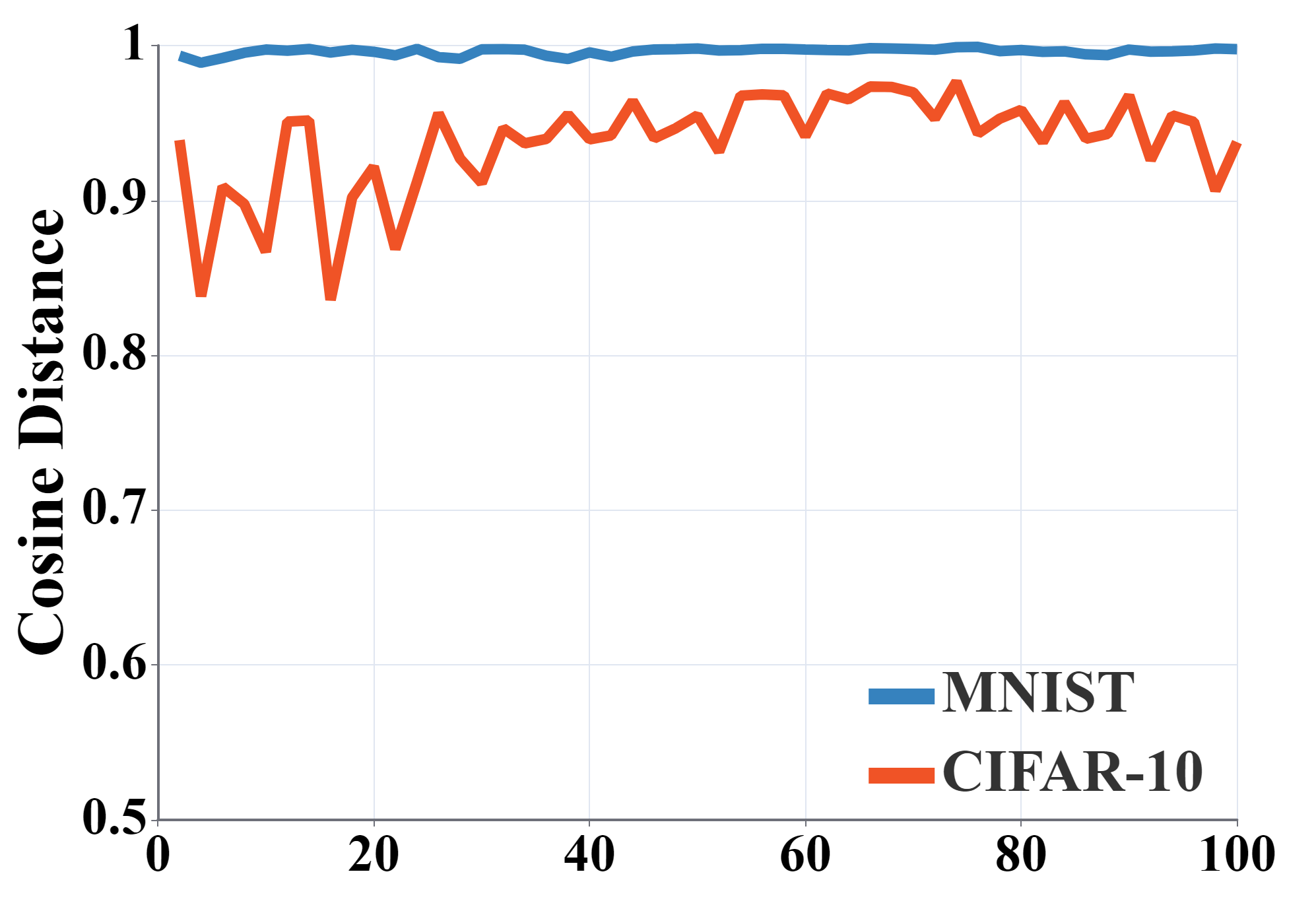}
    \caption{The cosine similarity between predictions of adversarial examples of identical examples in neighboring iterations.}
    \label{simi}
\end{figure}
Then, the direct application of Eq.~\ref{weight1} needs the defender to prepare adversarial examples for each $x_i\in D$ from the start to obtain $w^i$, which is still costly.
\sysname avoids the problem by utilizing the property that the model outputs similarly over adversarial examples crafted from the identical sample at adjacent iteration.
As illustrated in Fig. \ref{simi}, it can be observed that the adversarial examples crafted in previous iterations are also fairly comparable (in terms of ASR and model prediction) to the adversarial examples crafted in the current iteration.
Furthermore, it suggests that the importance of an example in the current iteration can be approximately evaluated by its previous adversarial versions, without employing its adversarial example.
Therefore, the sampling weight of the $i$-th training example in $t$-th iteration can be reformulated as follows:
\begin{equation}
\label{weight2}
w^i_t = \alpha w^i_{t-1} + (1-\alpha) w^i,
\end{equation}
where $w^i_t$ represents the weighted accumulation of threat intensity of previous adversarial examples of $i$-th training example and $alpha$ is the hyperparameter ($w^i_t = w^i_{t-1}$ if an example is not selected in $t$-th iteration).
Moreover, the higher the $\alpha$ is, the more attention the model pays to previous versions of adversarial examples.

\textbf{Practical Tricks.}
It is well known that NNs are usually trained based on the mini-batch stochastic gradient descent.
However, after applying the weighted sampling strategy mentioned above, a mini-batch of data can contain repeated samples (or samples with the same class), which can lower the learning efficiency of AT according to the active learning theory~\cite{sampling_strategy}.
To address the problem, we improve the sampling strategy used in \sysname from two aspects.
\par
First, \sysname adopts the strategy about sampling without replacement to avoid involving repeated samples.
Second, we introduce the class-wise balance strategy into \sysname to avoid that the samples used at each iteration belong to the same class.
In more details, the class-wise balance strategy evenly allocates the number of samples for each class during the sampling process.
Thus, every mini-batch of data is ensured to contain the samples of each class.

\section{Experiment}

\subsection{Experiment Setup}
We exhaustively examine the effectiveness of \sysname compared with vanilla AT~\cite{pgd_training} in two widely-used benchmark datasets, i.e., MNIST and CIFAR-10.
Specifically, we adopt ResNet18 as the base model throughout the experiments.
Moreover, for fair comparisons, we set the same training parameters for \sysname and AT.
For \sysname, we use $\alpha$ of 0.5.
To evaluate the performance of \sysname, we adopt natural accuracy (accuracy of natural samples), robust accuracy (accuracy of adversarial examples), and the running time for convergence as metrics.
The mini-batch size is set to 128 by default.
Referring to~\cite{free_training}, the adversarial examples are all crafted with PGD of 20 iterations and 10 restarts.

\begin{table}[h]
\centering
\caption{The required times (seconds) of AT and \sysname to achieve identical accuracy and robust accuracy in MNIST.}
\resizebox{0.23\textwidth}{!}{
\subtable[Accuracy]{
\begin{tabular}{@{}c|cc@{}}
\toprule
Accuracy (\%) & AT & \sysname \\ \midrule 
80 & 166.4 & 141.9 \\
85 & 192.0 & 167.7\\ 
90 & 243.2 & 219.3 \\
95 & 396.8 & 296.7 \\ \bottomrule
\end{tabular}
}
}
\resizebox{0.23\textwidth}{!}{
\subtable[Robust Accuracy]{
\begin{tabular}{@{}c|ccc@{}}
\toprule
\begin{tabular}[c]{@{}c@{}}Robust\\Accuracy (\%)\end{tabular} & AT & \sysname \\ \midrule
60 & 179.2 & 167.7 \\
70 & 230.4 & 219.3 \\
80 & 371.2 & 335.5 \\
90 & 1075.2 & 941.8 \\ \bottomrule
\end{tabular}
}
}
\label{table_mnist}
\end{table}

\begin{table}[h]
\centering
\caption{The required times (hours) of AT and \sysname to achieve identical accuracy and robust accuracy in CIFAR-10.}
\resizebox{0.23\textwidth}{!}{
\subtable[Accuracy]{
\begin{tabular}{@{}c|cc@{}}
\toprule
Accuracy (\%) & AT & \sysname \\ \midrule
65 & 1.33 & 0.57 \\
70 & 3.51 & 0.76 \\
75 & 7.31 & 1.52 \\
80 & 11.87 & 4.66 \\ \bottomrule
\end{tabular}
}
}
\resizebox{0.23\textwidth}{!}{
\subtable[Robust Accuracy]{
\begin{tabular}{@{}c|ccc@{}}
\toprule
\begin{tabular}[c]{@{}c@{}}Robust\\Accuracy (\%)\end{tabular} & AT & \sysname \\ \midrule
50 & 0.85 & 0.67 \\
55 & 2.09 & 1.33 \\
60 & 7.50 & 2.37 \\ \bottomrule
\end{tabular}
}
}
\label{table_cifar}
\end{table}

\subsection{Result}
\textbf{Overall results.}
First, we evaluate the overall performance of our method from the three main metrics, as reported in Table~\ref{table_mnist} and \ref{table_cifar}.
For MNIST and CIFAR-10, the accuracy and robust accuracy are evaluated at 1-iteration interval and 10-iteration intervals, respectively.
The baseline points in MNIST apply a bigger span because it is an easy-to-learn dataset.
From the results, \sysname achieves faster convergence speed in all settings, especially in the more sophisticated dataset CIFAR-10.
For instance, \sysname just requires 2.37 hours to reach 60\% robust accuracy; whereas, AT requires about triple times for convergence than \sysname in the same condition.
Likewise, in terms of accuracy, the times required to achieve different baseline points are considerably reduced (about faster 2$\sim$5x compared to AT).


\begin{figure*}
    \centering
    \subfigure[128]{\includegraphics[width=0.23\textwidth]{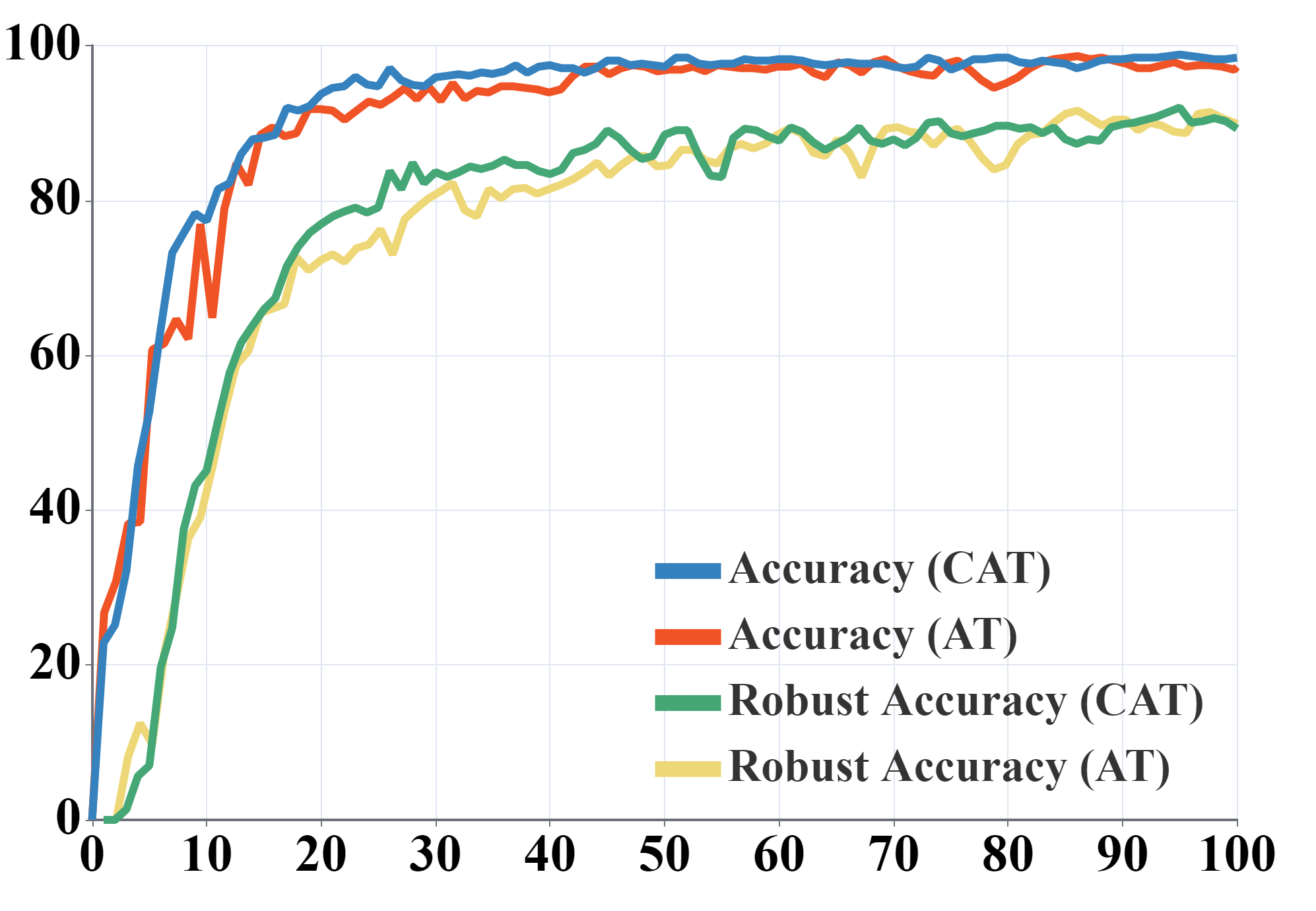}}
    \subfigure[256]{\includegraphics[width=0.23\textwidth]{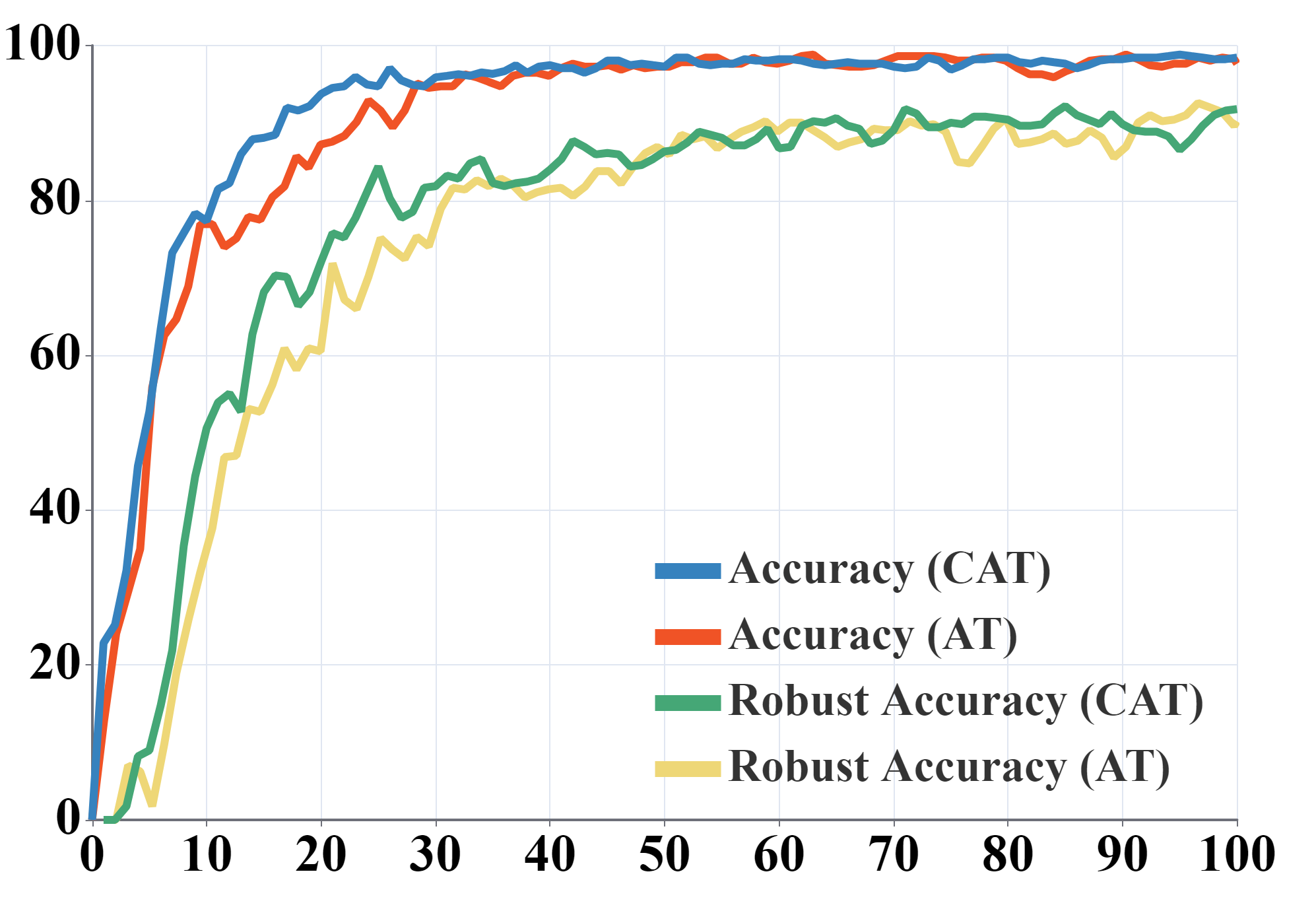}}
    \subfigure[512]{\includegraphics[width=0.23\textwidth]{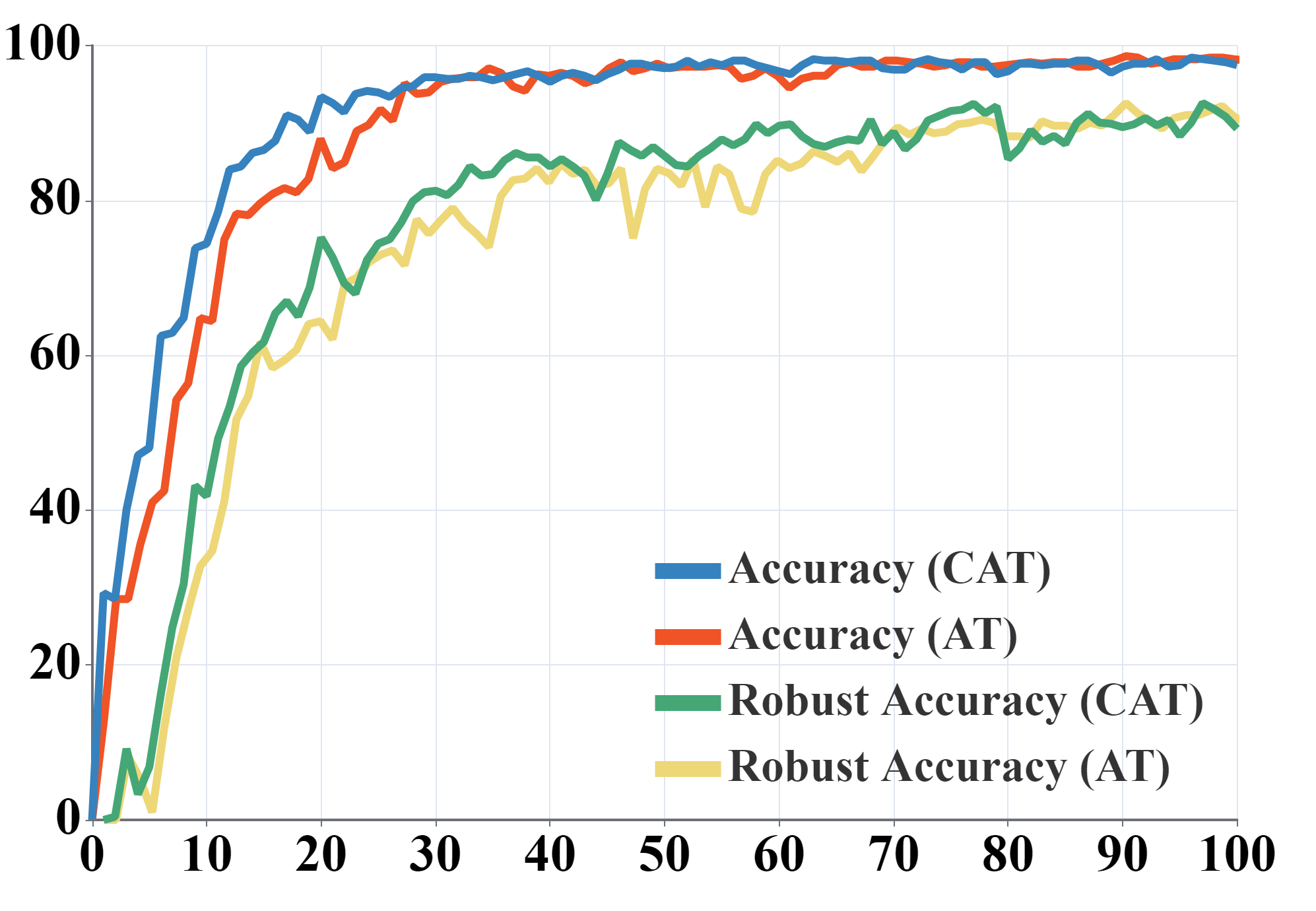}}
    \subfigure[1024]{\includegraphics[width=0.23\textwidth]{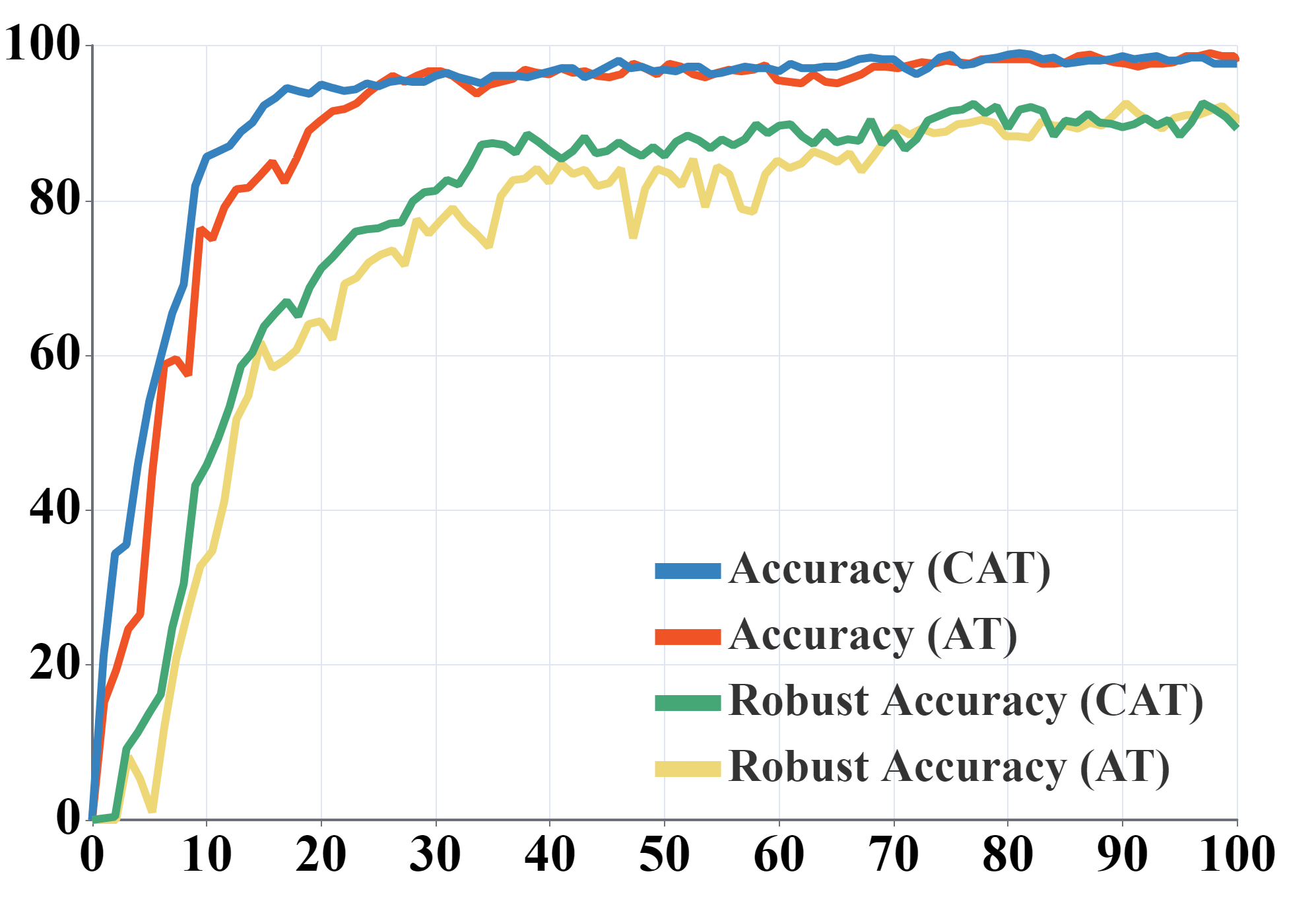}}
    \caption{The accuracy and robust accuracy of the model with AT and \sysname using different sampling numbers over different iterations in MNIST.}
    \label{trend_mnist}
\end{figure*}

\begin{figure*}
    \centering
    \subfigure[128]{\includegraphics[width=0.23\textwidth]{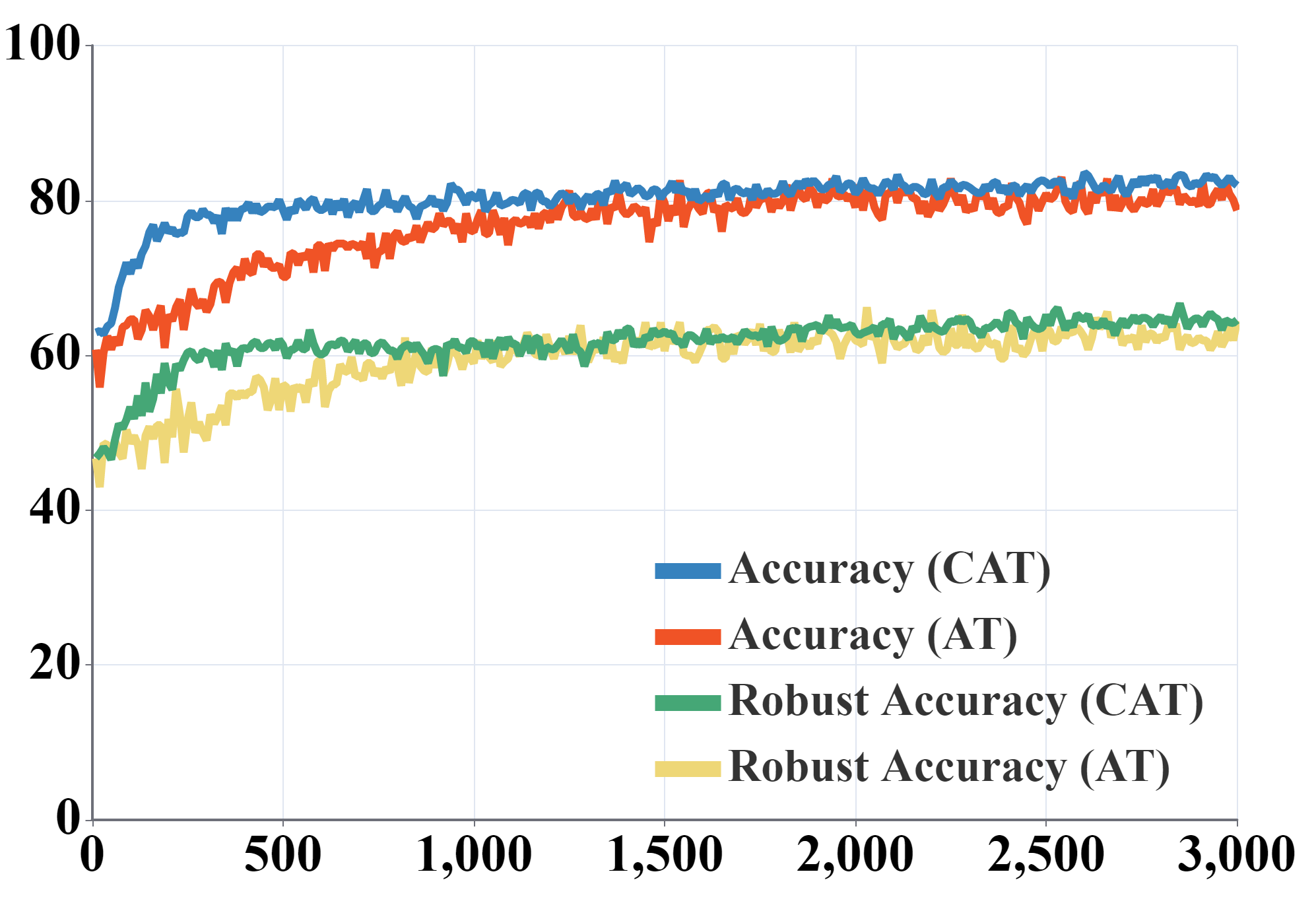}}
    \subfigure[256]{\includegraphics[width=0.23\textwidth]{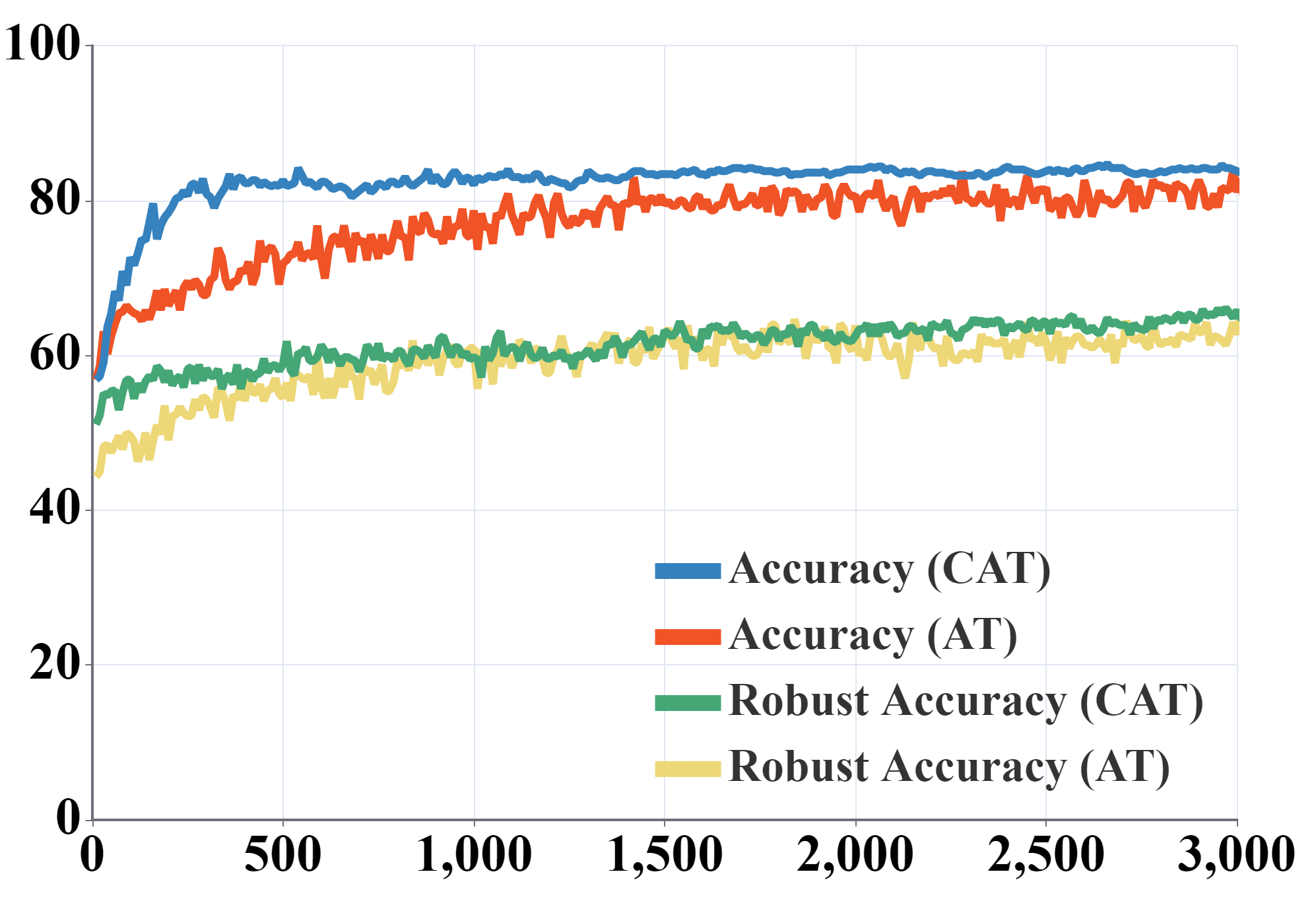}}
    \subfigure[512]{\includegraphics[width=0.23\textwidth]{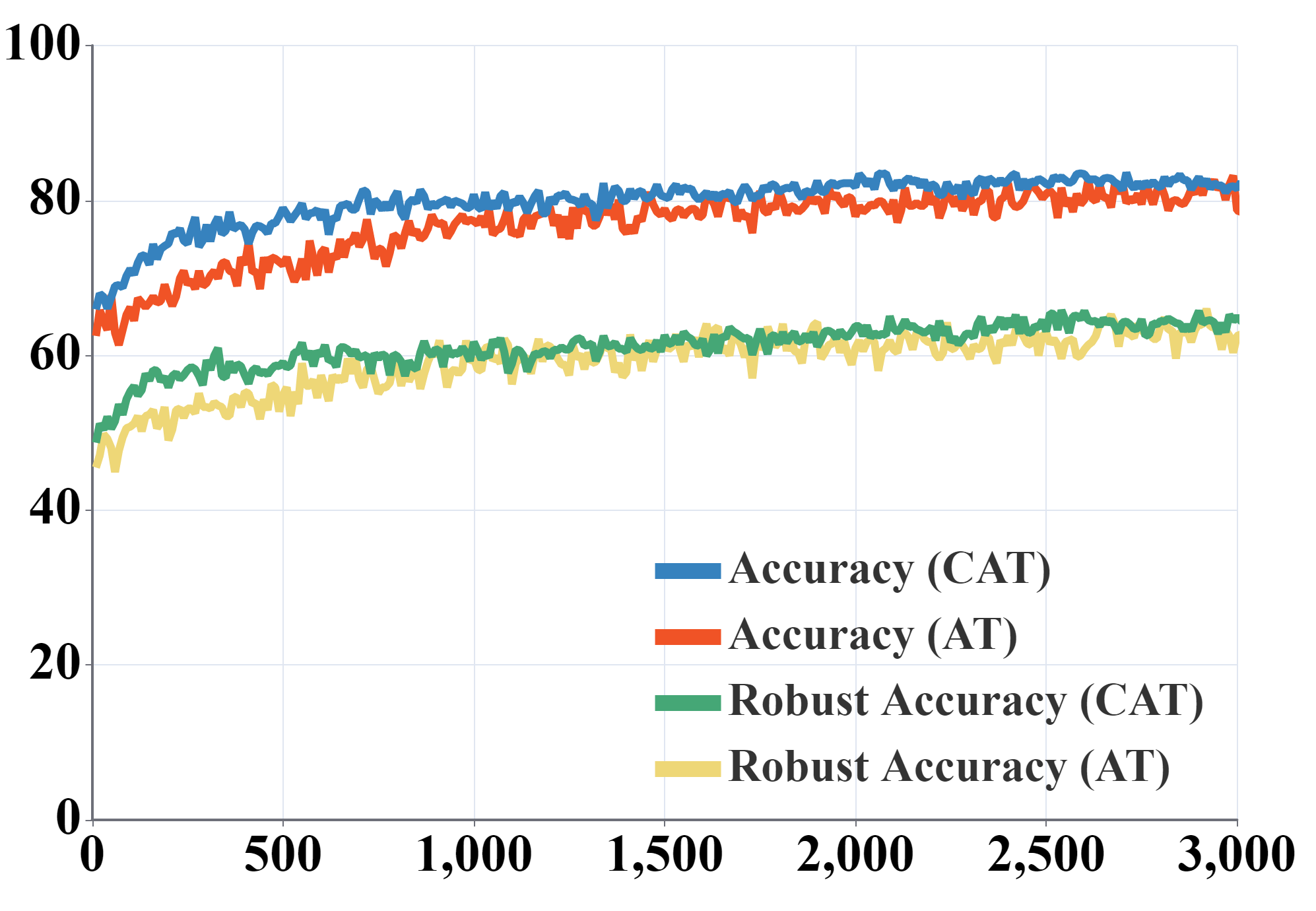}}
    \subfigure[1024]{\includegraphics[width=0.23\textwidth]{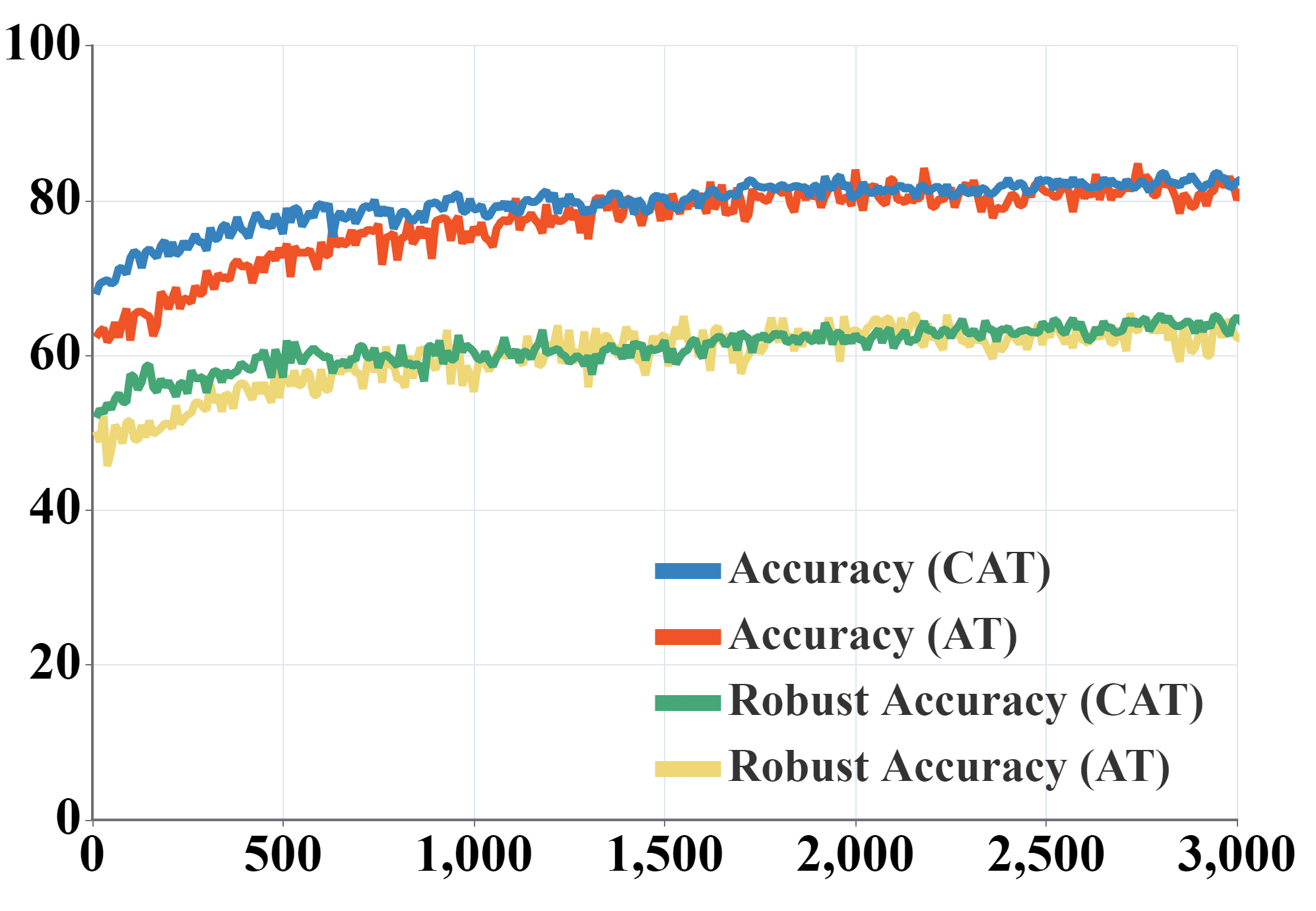}}
    \caption{The accuracy and robust accuracy of the model with AT and \sysname using different sampling numbers over different iterations in CIFAR-10.}
    \label{trend_cifar}
\end{figure*}

\noindent
\textbf{A closer look at \sysname.}
According to our design, a crucial factor that affects the performance of \sysname is the number of adversarial examples sampled at each iteration. 
Here, we examine the performance of \sysname over various sampling numbers in MNIST and CIFAR-10, as shown in Figure \ref{trend_mnist} and \ref{trend_cifar} (MNIST for 100 iterations and CIFAR-10 for 3000 iterations).
In the experiments, all hyperparameters are followed afore-mentioned ones.

First, from the experiments, we observe that as achieving identical accuracy or robust accuracy (i.e., the same defense effect), \sysname always requires fairly fewer iterations than AT for convergence.
The improvements mainly have the benefit of the sampling strategy used by \sysname, which can selectively filter informative samples to train the model, instead of equally treating each sample.
Moreover, the characteristic also leads to the fact that the speed-up effect of \sysname is more striking with the complicated learning task.
Thus, as shown in Figure \ref{trend_mnist} and \ref{trend_cifar}, the convergence speedup with CIFAR-10 is faster than MNIST.

Then, it can be discovered that there is a trade-off of the convergence speed and the sampling number for \sysname.
Specifically, the convergence speed can rise with increasing sampling number when the sampling number is below a certain threshold, while the increased sampling number instead weakens the efficiency of \sysname when the sampling number is higher than the threshold.
For example, in the case of CIFAR-10, the convergence speed rises in smaller sampling numbers until its peak of around 256, and then the convergence speed gradually decreases with the increased sampling numbers.
This phenomenon can be comprehended by considering two aspects.
On the one hand, if the sampling number is small, in each iteration, only a few samples are involved, i.e., only weights of a few samples can accurately reflect the actual gains brought to the model.
In other words, the weights based on adversarial examples crafted in previous iterations can no longer accurately (or approximately) measure the constructive information contained by the adversarial examples because of the excessive update frequency of the model.
On the other hand, as the sampling number remarkably increases, the performance of random sampling has to approach the weighted sampling gradually.
Loosely speaking, it can be intuitively regarded as that, the highly informative adversarial examples can be sufficiently sampled with the current sampling number; if the sampling numbers are increased, the additional adversarial examples sampled can only provide quite limited contribution to the model.

\section{Conclusion}
Inspired by the philosophy of active learning, we presented a novel method, i.e., case-aware adversarial training (\sysname) to improve the efficiency of adversarial training.
The core of the method was to relieve the inherent major flaw (high computation cost) of adversarial training by selecting samples with rich information at each iteration.
During the process, the weights of adversarial examples could not be accurately derived from the original examples.
To overcome the problem, we proposed a likelihood-based method to measure the information gain of adversarial examples.
Moreover, we also introduced two practical tricks to improve the diversity of each adversarial training mini-batch.
Extensive experiments
showed that \sysname could significantly accelerate the existing adversarial training method.
Finally, we highlighted that \sysname was a generic training scheme and it could be effortlessly combined with other adversarial methods.

\bibliographystyle{IEEEbib}
\bibliography{reference}

\end{document}